\def\year{2019}\relax
\documentclass[letterpaper]{article} 
\usepackage{aaai19}  
\usepackage{times}  
\usepackage{helvet}  
\usepackage{courier}  
\usepackage{url}  
\usepackage{graphicx}  
\usepackage[british]{babel}
\usepackage{hhline}
\usepackage{multirow}
\usepackage{amsmath,amssymb}
\usepackage[colorlinks]{hyperref}
\usepackage{subcaption}
\usepackage{float}
\usepackage[textsize=scriptsize,backgroundcolor=red!10,bordercolor=red!90]{todonotes}

\begin{document}
%
\title{Fair Transfer of Multiple Style Attributes in Text}
\author{
{K. Dabas}\textsuperscript{$\dagger$},
{N. Madaan}\textsuperscript{$\star$},
{G. Singh}\textsuperscript{$\star$},
{V. Arya}\textsuperscript{$\star$},
{S. Mehta}\textsuperscript{$\star$}, 
{T. Chakraborty}\textsuperscript{$\dagger$}\\
\textsuperscript{$\dagger$}IIIT Delhi, India\\
\textsuperscript{$\star$}IBM Research, India\\
\textsuperscript{$\dagger$}\{karan15141@iiitd.ac.in, chak.tanmoy.iit@gmail.com\}\\
\textsuperscript{$\star$}\{nishthamadaan, vijay.arya, sameepmehta\}@in.ibm.com, \{singh.gautam.iitg@gmail.com\}
}
\maketitle
\begin{abstract}
To preserve anonymity and obfuscate their identity on online platforms users may morph their text and portray themselves as a different gender or demographic. Similarly, a chatbot may need to customize its communication style to improve engagement with its audience. This manner of changing the style of written text has gained significant attention in recent years. Yet these past research works largely cater to the transfer of single style attributes. The disadvantage of focusing on a single style alone is that this often results in target text where other existing style attributes behave unpredictably or are unfairly dominated by the new style. To counteract this behavior, it would be nice to have a style transfer mechanism that can transfer or control multiple styles simultaneously and fairly. Through such an approach, one could obtain obfuscated or written text incorporated with a desired degree of multiple soft styles such as female-quality, politeness, or formalness. 

In this work, we demonstrate that the transfer of multiple styles cannot be achieved by sequentially performing multiple single-style transfers. This is because each single style-transfer step often reverses or dominates over the style incorporated by a previous transfer step. We then propose a neural network architecture for fairly transferring multiple style attributes in a given text. We test our architecture on the Yelp data set to demonstrate our superior performance as compared to existing one-style transfer steps performed in a sequence. 
\end{abstract}

\section{Introduction}
Style is an essential ingredient of effective communication, and by varying it one can convey more or less information than the literal meaning of words~\cite{Hovy-1988,rao2018dear}. Therefore reasoning about language style is a foundational capability needed to build successful NLP/G systems. For instance, a chatbot whose communication style or tone is tailored to its audience and context is more likely to trigger their engagement~\cite{hu2018touch}. Additionally, automated assistants capable of recognizing and modifying the style of text for emails, social media posts, online reviews, resumes, and marketing communications can be extremely valuable to both businesses and individual users. While NLP techniques have advanced significantly in extracting and understanding text content, altering the style of text is challenging due to a number of reasons. Firstly, both content and style are generally entangled together in text to convey specific information. Style impacts all aspects of text including punctuation, syntax, lexical choice, grammatical structure and layout. Therefore techniques must be designed to decouple style from content and alter it a manner that it still preserves the underlying content and generates syntactically correct and coherent text alternatives. Secondly, most style transfer problems lack parallel data(a dataset in which every data point belonging to one class has the corresponding data representation in other class. For example, a dataset of English to French sentences where each English sentence have a corresponding French representation in a form of a sentence will be called a parallel data), which makes this an unsupervised machine learning problem that is harder to solve. More recent neural network techniques such as variational auto-encoders (VAEs) and generative adversarial networks (GANs) that are designed to work with non-parallel vision datasets are not easily extensible due to the discrete nature of the text~\cite{shen2017style,corr-1805-11749}.

The goal of text or writing style transfer is to migrate the content of a sentence from a source style to a target style while still retaining the style-independent content. Most of the existing style transfer techniques have focussed on individual and granular styles; for example, altering the sentiment of a review from negative to positive style, transferring a tweet from male to female style, and converting the style of text from one author to other~\cite{li2018delete,reddy2016obfuscating,JhamtaniGHN17,rabinovich2016personalized}. More recent work has attempted to transfer individual soft styles such as impolite to polite and informal to formal~\cite{D16-1216,rao2018dear}. In practice however, we are often interested in incorporating multiple styles in the same content. For instance, a chatbot communicating with an elderly woman may have to simultaneously introduce multiple stylistic attributes such as `formal', `polite', and `female'. Similarly, to obfuscate her identity, an online user may be interested in simultaneously convert the review style from 'female' to 'male' and 'informal' to 'formal'. 

In this work, we formulate and study the \emph{multi-style transfer problem} with a focus on multiple soft styles that can fairly co-exist together within the same text. Our goal here is to transfer the content of a sentence from a set of source styles to a set of target styles while still preserving the meaning of the sentence. For instance, given a source sentence that holds `male' and `formal' stylistic attributes, we wish to convert it into target sentence with styles `female' and `informal' or `male' and `informal'. We study two approaches to solve this problem. We first use existing single style-transfer techniques to sequentially introduce multiple individual styles within a sample. We use different datasets and state of the art techniques to demonstrate that such an approach is ineffective as each new style interacts with the previously introduced styles in an unfair manner and either nullifies or dominates over previous styles. We then extend the work of~\cite{prabhumoye2018style} and design a novel architecture for multi-style transfer problem. As in prior work, the proposed architecture uses a back-translation model to rephrase the source sentence and convert it into its latent representation that is devoid of style specific features. 
However, in our work, this is then fed as input to a decoder that receives feedback jointly from multiple classifiers, each corresponding to a target style. We evaluate the effectiveness of the proposed approach(STMS: Style Transfer for Multiple Styles) for both content preservation and style strength accuracy using style classifiers trained on held-out data. We show that the proposed approach shows improvements and is more fair as compared to baseline approaches that introduce individual styles sequentially. The proposed model was able to convert a female sentence into male and formal with content preservation of 0.866 and style strength of 0.249 for female class, 0.751 for male class, 0.241 for formal class and 0.758 for informal class in comparison to source/input sentences which had style strength of 0.758 for female class, 0.242 for male class, 0.167 for formal class and 0.833 for informal class.

\section{Related Work}
\label{sec:related}
The style transfer problem has been addressed in the past but only for one style at a time. These past approaches can be split across two dimensions: 1) those that require parallel data, i.e. sentence pairs representing the same content in source and target styles, and 2) those that do not require parallel data, i.e., the data contains source style representing some content and target style representing another content. The first kind of data is much harder to obtain; hence most recent approaches make use of non-parallel data. We explain the detailed approaches taken for each of the cases.

\subsection{Approaches using parallel data}

\cite{hu2018touch} trained a sequence to sequence model using chat question-answer pairs. For each QA pair, tone label of the answer is given. This tone label is appended to the encoded question's vector so that decoder learns to generate an answer of the corresponding tone label.

\subsection{Approaches using non-parallel data}
The main intuition used in approaches dealing with non-parallel data is to sample a sentence from the target style such that its content is close to the source sentence. 

One of the early work by \cite{reddy2016obfuscating} performs a \emph{lexicon based} substitution on the words of the source sentence in order to transform its author style from male  to female. It identifies the words that are good candidates for substitution (say, 'bro'), finds appropriate substitutes from the target style (say, 'love') and performs substitution based on a `substitutability score'.

 \cite{li2018delete} proposed a method using \emph{phrase based} attribute markers in a source/target sentence. The method first deletes the source attribute markers from a given sentence, retrieves the target attributes and then generates the output sentence using RNN given the target style. Authors show an overall success rate of about 43\% on the input sentences. However the technique does not control the balance between content preservation and style-transfer. 

\cite{shen2017style} proposed a cross-aligned auto-encoder based architecture for style transfer task. The work focuses on converting sentiment from positive to negative or vice versa. The suggested technique performs well on reviews datasets, and the evaluation is done using a bidirectional LSTM based classifier. Further \cite{prabhumoye2018style} argued on using author's characteristics as pointed out by \cite{rabinovich2016personalized} who  argued that in a machine translation system, author's characteristics are very well obfuscated. Hence \cite{prabhumoye2018style} worked on building a feedback based machine translation system that tends to obfuscate author's writing style to mask the characteristics of author's writing. Then, to do the style transfer the system learns a latent representation of the input sentence in a language translation model to preserve the content and hide the style. Following this, they employed an adversarial generation technique to match the output with the desired one. 


 A closely related problem which plays also a significant role in style transfer problems is bias in the data. We review prior literature related to this in the following subsections.

\noindent\textbf{De-biasing the training algorithm} as a way to remove the biases focuses on training paradigms that would result in fair predictions by an ML model. In the Bayesian network setting, Kushner et al. proposed a latent-variable based approach to ensure counter-factual fairness in ML predictions. 

\noindent\textbf{De-biasing the model after training} as a way to remove bias focuses on ``fixing" the model after training is complete. \cite{bolukbasi2016man} in their famous work on gender bias in word embeddings took this approach to ``fix" the embeddings after training.

\noindent\textbf{De-biasing the data at the source} fixes the data set before it is consumed for training. This is the approach we take in this paper by trying to de-bias the data by converting text into multiple styles as per user's input. A related task is to modify or paraphrase text data to obfuscate gender as in \cite{reddy2016obfuscating}. Another closely related work is to change the style of the text to different levels of formality as in \cite{rao2018dear}.


\section{Problem Formulation}
\label{sec:problem}
As mentioned before, our goal is to morph the style of the text along more than one style attributes. Each style attribute furnishes two styles. For example, the gender style attribute has a male style and a female style. Therefore, if there are $n$ style attributes, we would have sample sentences of $2n$ styles. 

In this paper, we deal with the case of simultaneous transfer of two style attributes. Therefore, if our two style attributes are \emph{gender} and \emph{formality}, then we would need two data sets - the first containing samples of male and female sentences and the second containing samples of formal and informal sentences. Let the two style attributes be denoted by $s_1$ and $s_2$, and let the corresponding data sets be denoted by $D_1$ and $D_2$. 

We also assume the presence of an arbiter function $c(x)$ that consumes a sentence $x$ and returns a style-tuple $(a_1, a_2)$, where $a_1$ and $a_2$ are scores between 0 and 1 and represent the strengths the two styles $s_1$ and $s_2$ of the sentence. In our case, $a_1$ refers to the maleness of the author style of the sentence, and $a_2$ refers to the degree of formality in the given sentence.

Given any input sentence $x$ and given any target style-tuple $(1,1)$, $(1,0)$, $(0,1)$ or $(0,0)$ - the goal of our model is to return a new sentence $y$ such that $c(y)$ is close to the target style-tuple, and the content of $y$ is close to the content of the original sentence $x$. 



\section{Methodology}
\label{sec:method}
Let D\textsubscript{1} = \texttt{\{}s\textsubscript{11}\textsuperscript{1}, ...,s\textsubscript{11}\textsuperscript{n}, s\textsubscript{12}\textsuperscript{1}, ...,s\textsubscript{12}\textsuperscript{n}\texttt{\}} and D\textsubscript{2} = \texttt{\{}s\textsubscript{21}\textsuperscript{1}, ...,s\textsubscript{21}\textsuperscript{n},  s\textsubscript{22}\textsuperscript{1}, ...,s\textsubscript{22}\textsuperscript{n}\texttt{\}} be two different datasets, each containing a different style attribute (e.g. Gender, Formality). The objective is to morph the style along different style attributes while still maintaining the content of the input sentence. Assuming that we want to add styles s\textsubscript{11} and s\textsubscript{21} in the input sentence. We represent this operation as S\textsubscript{1\&2}\textsuperscript{1}, where the symbol `\&' denotes the fusion of styles into the input sentences.

\begin{figure}
\centering
\includegraphics[width=\columnwidth, height=4cm]{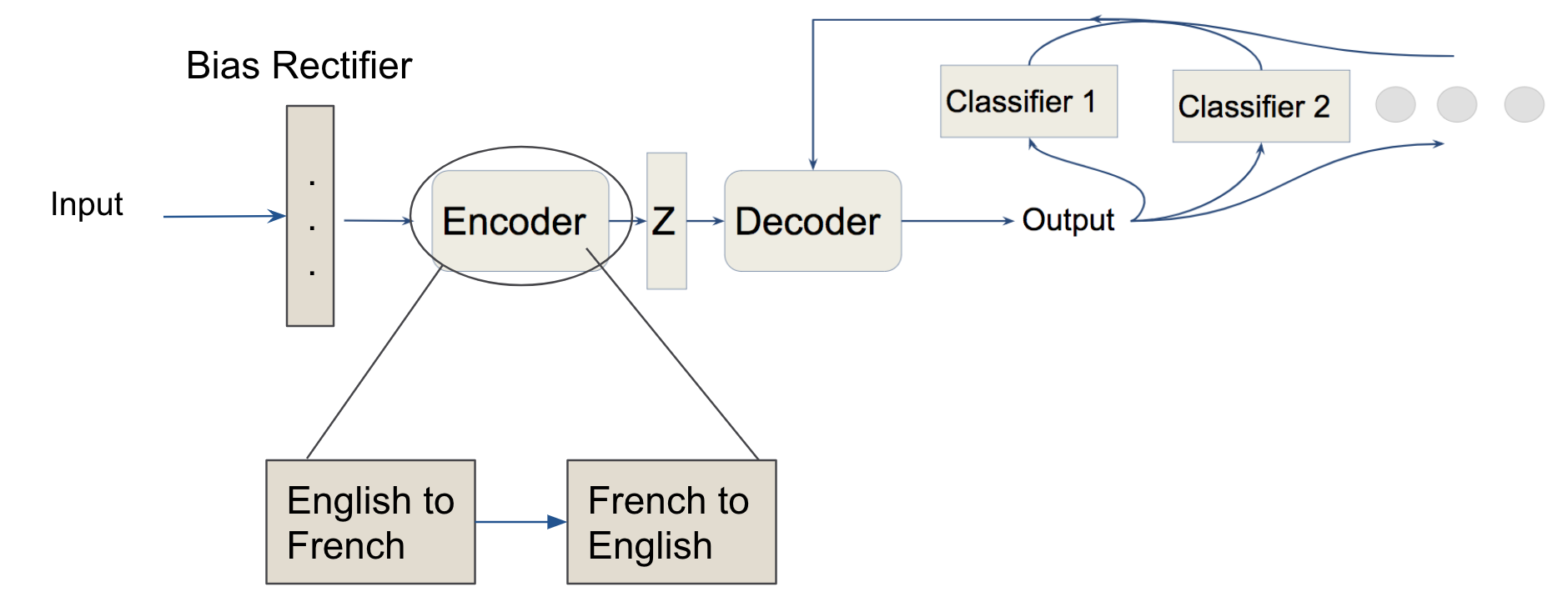}
\caption{STMS Model: A Feedback based Architecture for multi-Style Transfer problem}
\label{fig:Exp2}
\end{figure}

Our the model contains a latent content variable Z such that it only has the meaning of the input sentence and is free of any bias from author’s stylistic features. They achieved this by translating the input sentence into a foreign language (in this case to French) and then translating it back to the source language (in this case English). Such translation tends to normalize the input sentences and makes it free of author’s stylistic features. We use this normalization techniques in our STMS(Style Transfer for Multiple Styles) model. We use a natural machine translation model which is capable of converting the English sentences to French followed by a back translation model which translate the French sentences to English. Figure \ref{fig:Exp2} shows the overview of our STMS model. We begin with the sentence style normalization in the BST model \cite{prabhumoye2018style}. Following this, we train the single decoder for multiple style fusion/combining into the output sentence. We use N (in this case N = 2) style classifiers in a feedback loop to the decoder. The output sentences from our model will have both the styles S\textsubscript{1\&2}\textsuperscript{1}.
\begin{table}
\begin{center}
 \begin{tabular}{|c|c|} 
 \hline
 \textbf{Class} & \textbf{\# of sentences(Million)}\\
 \hline
 Male & 2.892\\ [0.25ex]
 \hline
 Female & 2.892\\[0.25ex]
 \hline
\end{tabular}
\caption {Statistics for Gender Dataset}
\label{Yelpdata}
\end{center}
\end{table}

\subsection{Bias Rectifier}
Bias rectifier is an initial layer of filter used to balance the dataset for each attribute. In the dataset section, we observe that the Gender dataset is heavily biased as most of the sentences present in either of the classes (i.e. Male, Female) are informal sentences. We use a pre-trained model for getting the formal/informal labels on the Gender dataset, and if the majority of sentences fall under one class (e.g. informal), we try to balance the classes by duplicating the sentences in the minority class. This type of rectification helps in making the learning free from any bias present in the dataset.  
\begin{table}
\begin{center}
 \begin{tabular}{|c|c|} 
 \hline
 \textbf{Class} & \textbf{\# of sentences(Thousand)}\\
 \hline
 Formal & 108.682\\ [0.25ex]
 \hline
 Informal & 108.682\\[0.25ex]
 \hline
\end{tabular}
\caption {Statistics for Formality Dataset}
\label{Formaldata}
\end{center}
\end{table}

\subsection{Style Specific Classifiers}
Our STMS model uses multiple style classifiers to provide a feedback to the proposed models. The classifier is a convolutional neural network (CNN) which classifies a sentence into multiple style categories (e.g. Male, Female, Formal and Informal). We trained style classifiers for each style attribute  (N = 2). These classifiers are trained in a supervised fashion. Assuming that the style attribute is S =\texttt{\{}s\textsubscript{1},s\textsubscript{2}\texttt{\}}, since each style attribute has two categories (e.g. Gender attribute has categories as Male\texttt{\{}s\textsubscript{1}\texttt{\}}, Female\texttt{\{}s\textsubscript{2}\texttt{\}}). Let D denote the dataset which holds the sentences labeled into these categories. The loss function for a single classifier is: 
\begin{equation}
\ell_{class}(\varnothing_{c}) =\mathop{\mathbb{E}}\textsubscript{D}[\log{qc(s|d)}]
\end{equation}
Where \( \varnothing_{c} \) denotes the classifier parameter, s is the style and d is the sentence in a class.

\subsection{Feedback-based Style Generation}
Figure \ref{fig:Exp2} shows that the decoder model is trained on latent content variable z which only contains the content representation of the sentences. Similar to  \cite{prabhumoye2018style} we train the decoder model with few major differences such as (i) We only train the single decoder in comparison to multiple decoder (one for each style), (ii) The series of style classifiers (N = 2, i.e., two style attribute classifiers) are used as a feedback system in training a single decoder which can induce multiple styles in the generated output. The decoder model consists of bi-directional LSTM which is used to generate the output sequence. The encoder model creates a latent content variable z which is defined as:
\begin{equation}
z = Enc(d)
\end{equation}
The loss function defined for the decoder is:
\begin{equation}
\ell_{decoder}(\varnothing_{decoder\;d}) =\mathop{\mathbb{E}}\textsubscript{Enc(d)}[\log{p_{gen}(d|z)}]
\end{equation}

The loss is calculated using cross-entropy, where \(\varnothing_{decoder}\) denotes the decoder parameter. Since we are using a single decoder with a series of classifiers (N = 2) in a feedback to  the decoder, we need to share the classifier learning with the decoder in a form of a loss minimization problem. Therefore we do a weighted sum of the individual losses from decoder, classifier for style attribute-1, classifier for style attribute-2. So the final generative loss function is:
\begin{equation}
min_{\varnothing_{decoder}}\ell_{gen}  = \ell_{decoder}+ \alpha*\ell_{class1} + \beta*\ell_{class2}
\end{equation}

Where \(\alpha\) and \(\beta\) are the balancing parameters for the two classifiers, \(\ell_{decoder}\) is the decoder loss, \(\ell_{class1}\) is the classifier loss of first style attribute and \(\ell_{class2}\) is the classifier loss of second style attribute.

\section{Experiments and Results}
\label{sec:exp}
In this section we show the baseline models used for comparison, dataset statistics, evaluation techniques and discuss the results obtained from our STMS model and other baseline models.

\subsection{Baseline Models}
We use seq2seq model \cite{Sutskever2014SequenceTS}, cross-aligned auto-encoder model \cite{shen2017style} and back translation for style transfer model \cite{prabhumoye2018style} to draw the comparison with our proposed STMS(Style Transfer for Multiple Styles) model. Since the problem is to combine multiple styles into one sentence, as stated in the previous sections; existing baseline models were built for one-to-one style transfer so as to use these baseline models to work with multiple styles. Furthermore, we introduce the variants of these baseline models where we use a sequential approach to connect them. Let us assume that we have two sets of styles \texttt{\{}S\textsubscript{1},S\textsubscript{2}\texttt{\}} and \texttt{\{}S\textsubscript{3},S\textsubscript{4}\texttt{\}}, and the existing baseline models are trained to convert the style S\textsubscript{1} to S\textsubscript{2}, S\textsubscript{3} to S\textsubscript{4} or vice versa. Let us say we have 3 models seq2seq, CAE, BST models, representing seq2seq model, cross-aligned auto-encoder and back translation for style transfer model, respectively. In order to get a sentence with combinations of multiple styles, we first pass an input sentence to seq2seq/CAE/BST \texttt{\{}S\textsubscript{1},S\textsubscript{2}\texttt{\}} model and then the output from this conversion is passed to seq2seq/CAE/BST \texttt{\{}S\textsubscript{3},S\textsubscript{4}\texttt{\}} model to get the final output sentence which should have styles S\textsubscript{2}, S\textsubscript{4} in it. As a consequence, we have 5 baseline models -- seq2seq, CAE, BST, Sequential seq2seq, and Sequential CAE.

\subsection{Dataset Description}
We use two different datasets where one contains the sentences labeled as male/female (i.e., Gender dataset) and other set of sentences labeled as formal/informal(i.e. Formality dataset). The Gender dataset consists of publicly available Yelp reviews database which were labeled into male/female class based on the usernames of the reviewers. This labeling was done by~\cite{reddy2016obfuscating}. On the other hand, the Formality dataset was part of the Yahoo Answers corpus L6, which was labeled in~\cite{rao2018dear}. Table \ref{Yelpdata} shows that the Gender dataset  consists of 2.889M sentences per class. The dataset was split into train, test, dev and classifier set per class with 1.334M, 267.2K, 2.246K, 1.288M samples respectively. From Table \ref{Formaldata} we can see that the Formality dataset which is a part of GYAFC dataset \cite{rao2018dear} consists of 81.844K sentences per class. Out of 81.844K sentences the dataset was split into train, test, dev and classifier set per class with 51.967K, 1.332K, 2.788K, 52.595K samples respectively. Apart from this, we also conduct basis data analysis (see Figure \ref{fig:pun}) where we observe that the female class tends to use exclamation mark more often in their reviews compared to male class which uses a full stop at the end of a sentence. We use the classifier split from the above datasets to train our evaluation classifiers which are used in measuring the style strength of the model output. We also use these evaluation classifiers to find the style strength on the train split subset (which is used to train the style transfer models). We find (see Figure \ref{fig:pie}) that  88.1\% of sentences present in Gender dataset are informal in nature. This shows that if any style transfer model is trained on this type of dataset for male to female or female to male conversion, the output of such a model will always add a bias in the sentence in a form of informality.    


\begin{table}
  \centering
  \renewcommand{\arraystretch}{1.2}
  \begin{tabular}{|p{2.5cm}|c|c|c|c|}
    \hline
    \multirow{2}{3cm}{\textbf{Test sentences}} & \multicolumn{4}{c|}{\textbf{Style Strength}}\\
    \cline{2-5}
    & \textbf{Female} & \textbf{Male} & \textbf{Formal} &\textbf{Informal}\\
    \hline
    Female  & 0.758 & 0.242 & 0.167 & 0.833\\ \hline
   	Male  & 0.257 & 0.742 & 0.169 & 0.831\\ \hline
    \end{tabular}
  \caption{Style strength of Input test sentences}
  \label{test_set}
\end{table}

\subsection{Evaluation Techniques}
We evaluate the output sentences at two levels: (i) Content preservation (ii) Style strength. The main objective is to get a better style strength with high content preservation. But later we will see that the content preservation and style strength are inversely related. It means that one cannot get higher style strength without losing the content of the input sentence or vice-versa.
\begin{table}[h]
\begin{center}
 \begin{tabular}{|p{1.5cm}|p{3.7cm}|p{2cm}|} 
 \hline
 \textbf{Model} & \textbf{Conversion}& \textbf{Content Preservation}\\
 \hline
 seq2seq &Female to Male &\textbf{0.951}\\ [0.20ex]
 \hline
 CAE & Female to Male & 0.853 \\[0.20ex]
 \hline
 BST &Female to Male & 0.908 \\[0.20ex]
 \hline
 Sequential seq2seq & Female to Male\&Formal & 0.937\\[0.20ex]
 \hline
 Sequential CAE & Female to Male\&Formal & 0.626 \\  [0.20ex] 
 \hline
 STMS & Female to Male\&Formal &0.866\\ [0.20ex]
 \hline
\end{tabular}
\caption {Content preservation(0-1) and style transfer conversion attempted for each model}
\label{contentPre}
\end{center}
\end{table}

\begin{enumerate}
\item {\bf Content Preservation:} In order to check how well the content of the input sentences are preserved after the style transfer we find the preservation score which is defined as the cosine similarity between the input sentence vector and output sentence vector. We use a pre-trained glove embedding \cite{Pennington2014GloveGV} of 100 dimensions. The sentence vectors are formed as follows: Consider a sentence SS which consists of n words such as SS = \texttt{\{}w\textsubscript{1}, ...,w\textsubscript{n}\texttt{\}}. We find the word vectors for all n words in a sentence and simply take the mean of these n 100x1 dimension vectors. This gives us one 100x1 dimension vector which is used for cosine similarity between input sentence and output sentence.

\item {\bf Style Strength:} Apart from maintaining the content before and after style transfer we also want to know how well the target styles are infused in the output sentences. To measure the style strength we train two bi-directional LSTM classifiers one for each style attribute (i.e. Gender, Formality). The classifier has an accuracy of 81.03\% for male/female classes, and 79.7\% for formal/informal classes. These classifiers are used to evaluate the output sentences by giving the class score to it for these four classes(i.e. Male, Female, Formal, Informal).
\end{enumerate}

\begin{table}[h]
  \centering
  \renewcommand{\arraystretch}{1.2}
  \begin{tabular}{|p{2.3cm}|c|c|c|c|}
    \hline
    \multirow{2}{3cm}{\textbf{Model}} & \multicolumn{4}{c|}{\textbf{Style Strength}}\\
    \cline{2-5}
    & \textbf{Female} & \textbf{Male} & \textbf{Formal} &\textbf{Informal}\\
    \hline
    seq2seq & 0.592 & 0.408 & 0.165 & 0.835\\ \hline
    CAE & 0.527 & 0.473 & 0.164 & 0.836\\ \hline
    BST & 0.498 & 0.502 & 0.102 & 0.898\\ \hline
    Sequential seq2seq & 0.552 & 0.448 & \textbf{0.357}
 & \textbf{0.643}\\ \hline
    Sequential CAE & 0.377 & 0.622 & 0.130 & 0.870\\ \hline
    STMS & \textbf{0.249} & \textbf{0.751} & 0.241 & 0.758\\ \hline
  \end{tabular}
  \caption{Style transfer Strength after conversion (see Table \ref{contentPre} for types of conversion) for various models }
  \label{styleStrenght}
\end{table}
\subsection{Results}
We first measure the style strength on the test set. From Table \ref{test_set} we can say that the test set contains the sentences which are mostly informal in nature. And as we saw in the dataset section, the training data is also highly informal. We evaluate the models for converting the female sentences into the male and formal sentences. Given female input sentences the desired output should be male and formal in nature. Here the ideal result would be increase in male style attribute, increase in formal style attribute, decrease in female style attribute and decrease in informal style attribute. From Table \ref{contentPre} we can say that the content preservation is high in seq2seq model with a preservation score of 0.951 followed by sequential seq2seq, BST and STMS model with a preservation score of 0.937, 0.908, 0.866 respectively. The content preservation is least in sequential CAE model with a preservation score of 0.626. From Table \ref{styleStrenght} we see that the style strength is highest for gender attribute in STMS model where the Female style score is 0.249 and male style score of 0.751. And the Sequential seq2seq model gives the highest score for formal style in the output sentences with a score of 0.357 and STMS model gives a formal score of 0.241 which is the second highest. Overall the STMS model has given a better score compared to any other models as it increased both male style attributes and formal style attributes in the generated output.
We show some examples of generated sentences from the models stated in Table \ref{styleStrenght} where the input sentences were female and informal in nature and the objective is to get sentences which are both male and formal in nature(see Figure \ref{fig:Results_fig}).

\section{User Study}
\label{sec:userstudy}
To qualitatively assess the performance of our system, we conduct a user study, presenting the original and translated sentence to human users.
 Our user study has two phases.
 The first phase comprises of a set of sentences which are originally written by females and are informal(the reason behind choosing informal sentences is that mostly sentences written by females are informal), are shown to the reviewers.
 The second phase had questions that are asked after the translated sentences are generated using our system, are also presented to the subject.

 \subsection{Nature of Dataset}
 \label{subsec:userstudy:nature}
\begin{figure}
\includegraphics[width=80mm,scale=0.8, height=9cm]{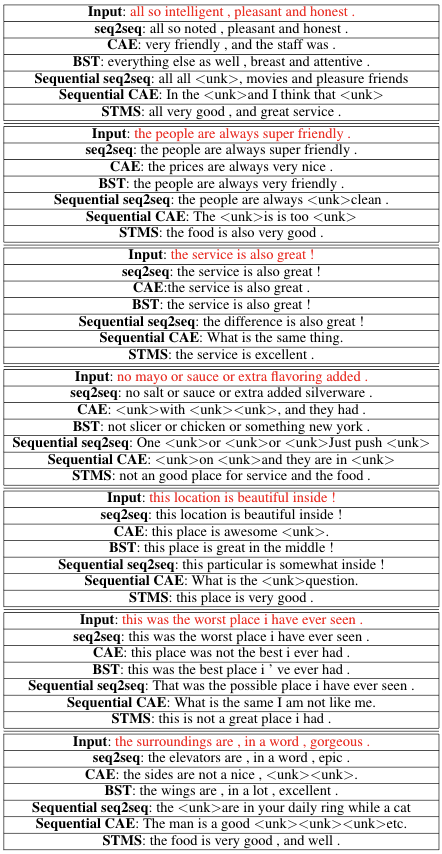}
\caption{Sample of Style translation of input sentence(Female) into Male, Formal style on Gender Dataset}
\label{fig:Results_fig}
\end{figure}

An example of input sentence can be looked for in Figure \ref{fig:Results_fig}

 \subsection{User Perception of Author Writing Bias}
 \label{subsec:userstudy:first}

 In the first phase, we qualitatively assess how writing style instills bias in the reviewer and how it affects people.
 We curated 10 reviews - female and formal in nature. Following each review, a question was asked that mapped the audience's understanding about the presence of bias from the in the review write-up. The questions consisted of if while reading the review you thought of gender of the author and then perceived it in that way?

 The third and fourth questions assess how well people can spot the bias in the written review by looking at the keywords used in the review.
 The study was conducted on a group of 15 individuals, aged (20-45), who were either graduates or graduate-level students of humanities (literature), and have worked in the past on raising awareness about gender bias in different spheres of society.
 8 males and 7 females participated.
 Both the genders were well-distributed over the age range.

 \subsection{Inferences and How it Motivates Our System}
 \label{subsec:userstudy:inferences}

 In majority of reviews, the participants both men and women felt that the author's gender and way of writing have a very high impact on how they take the review.

This also shows that there is a need to mask the gender and specific attributes of their writing before they are put outside from the perspective of gender bias, so that the reviews do not end up in reinforcing stereotypes about different genders in the style of writing.
 Hence we see our technology being used for social good, so that bias can be spotted and handled more efficiently through our system. 

 \subsection{User Evaluation of Our System}
 \label{subsec:userstudy:evaluation}
In the second phase, we performed a follow up user study on same subjects who took the above survey to see how the proposed system helps in mitigating the bias spotted by the subjects so that writing does not affect how the review is taken.
 Here, we presented the new translated version of reviews to the same users, and asked the following questions.

1) Were you satisfied with reading the review sentence generated by STMS?
2) Which are the highest content preserved sentences compared to source sentences? 
3) Which sentences do you feel are more formal in nature?
4) Which sentences do you feel reflect maleness and formal-ness in the given set of sentences?
5) Did you feel better after reading this new version of the review? 

Overall 15 of them said that the sentences generated from seq2seq model.
(ii) Which sentences do you feel are more formal in nature? 9 of them said that the sentences generated from the STMS model and 4 for BST and 2 for CAE.
(iii) Which sentences do you feel reflect maleness and formalness in the given set of sentences? Given that the source sentences have femaleness in them. 11 of them said that the sentences from STMS reflects more maleness and also have a level of formalness in them whereas, 4 of them believe that the BST model has the sentences with such styles.
 
 ~
Almost ~73\% of the people (11 people) were satisfied with the translation done by our system STMS. Overall 100\% of the people said that thee sentences generated by seq2seq model have highest content preservation. But when it came to formalness majority of them argued that STMS does a better job. 
~65\% of the users said that after reading the translated review they felt better.

\section{Conclusion and Future Work}
\label{sec:conclusion}
In this work, we introduced and studied the multi-style transfer problem where the goal is to translate a source text into a target text with multiple stylistic features while preserving the underlying meaning. We proposed two approaches to address this problem based on combining multiple individual style models in series and an extended encoder decoder architecture based on back-translation that utilizes feedback from multiple style classifiers to jointly guide a decoder. We used two datasets related to gender and formality and conducted multiple experiments to study the performance of existing state of the art classifiers for both individual as well as multi-style transfer problem using the proposed approaches. Our results indicate both Sequential seq2seq and STMS models improve the target style strength while retaining the underlying content. We further evaluate our results using a user study. 

A number of challenges need be addressed to successfully incorporate multiple stylistic features with high fidelity within a text sample. Firstly, we observe that mutual bias observed in datasets corresponding to each style can play a significant role in determining the content and style strength of the final outcome. In our experiments, we observed that most sentences in the gender dataset with male and female styles were informal in nature. As a consequence, female style when converted to male and formal invariably results in sentences that retain high informal style. Secondly, while designing style models to introduce multiple styles, multiple classifiers need to work in tandem and possibly compete against each other to provide feedback to guide the decoder. Therefore the weights assigned to loss functions of each classifier plays a significant role in impacting the output. These weights need to take into account the mutual bias between datasets. Also, since datasets corresponding to each style are non-parallel in nature without one-to-one correspondence, building classifiers trained on one dataset that work accurately on other datasets is a challenge. Lastly, existing style models that introduce one style at time seem to erase other styles within the sentence in order to introduce a style. This impacts approaches that combine multiple style models in series. As a consequence, individual style models that carefully erase only relevant styles while introducing a new style will be important. We plan to address some of these challenges in future work.

 \bibliographystyle{aaai}
 \bibliography{ref.bib}
\end{document}